\documentclass[10pt,twocolumn,letterpaper]{article}

\usepackage{wacv}              

\usepackage{graphicx}
\usepackage{amsmath}
\usepackage{amssymb}
\usepackage{booktabs}
\usepackage{subcaption}
\usepackage{array}
\usepackage{colortbl}
\usepackage{xcolor}
\usepackage{dblfloatfix}
\usepackage[normalem]{ulem}
\useunder{\uline}{\ul}{}

\usepackage[pagebackref,breaklinks,colorlinks]{hyperref}

\usepackage[capitalize]{cleveref}
\crefname{section}{Sec.}{Secs.}
\Crefname{section}{Section}{Sections}
\Crefname{table}{Table}{Tables}
\crefname{table}{Tab.}{Tabs.}

\def\wacvPaperID{
1981} 
\def\confName{WACV}
\def\confYear{2025}

\begin{document}

\title{PRoGS: Progressive Rendering of Gaussian Splats}

\author{Brent Zoomers,  Maarten Wijnants,  Ivan Molenaers,  Joni Vanherck,  Jeroen Put,  Lode Jorissen,  Nick Michiels\\
Hasselt University - Flanders Make - Expertise Centre for Digital Media, Diepenbeek, Belgium \\
{\tt\small{firstname}.{lastname}@uhasselt.be}
}

\twocolumn[{
    \maketitle
    \begin{center}
    \captionsetup{type=figure}
    \includegraphics[width=1\textwidth]{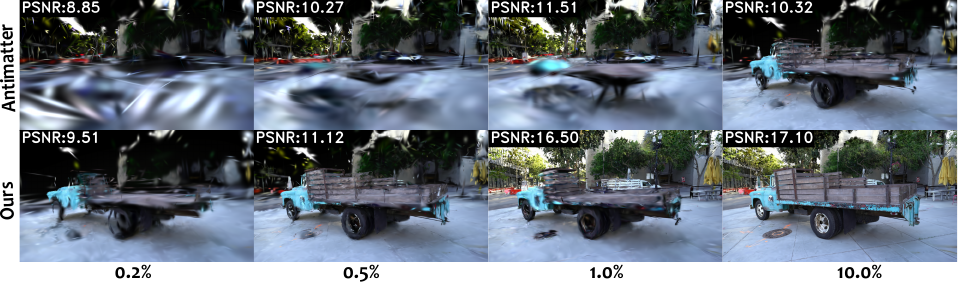}
    \captionof{figure}{Progressive Rendering of 3DGS using our approach versus using an existing web-viewer by Antimatter. From left to right, we show 0.2\%, 0.5\%, 1\%, and 10\% of the total number of splats, respectively. Our approach results in a faster visualization of a representative version of the scene. At 0.2\%, we have loaded in a basic level of the truck while it is completely missing in the alternative approach. }
    \label{initfigure}
\end{center}
}]

\begin{abstract}
Over the past year, 3D Gaussian Splatting (3DGS) has received significant attention for its ability to represent 3D scenes in a perceptually accurate manner. However, it can require a substantial amount of storage since each splat's individual data must be stored. While compression techniques offer a potential solution by reducing the memory footprint, they still necessitate retrieving the entire scene before any part of it can be rendered. In this work, we introduce a novel approach for progressively rendering such scenes, aiming to display visible content that closely approximates the final scene as early as possible without loading the entire scene into memory. This approach benefits both on-device rendering applications limited by memory constraints and streaming applications where minimal bandwidth usage is preferred. To achieve this, we approximate the contribution of each Gaussian to the final scene and construct an order of prioritization on their inclusion in the rendering process. Additionally, we demonstrate that our approach can be combined with existing compression methods to progressively render (and stream) 3DGS scenes, optimizing bandwidth usage by focusing on the most important splats within a scene. Overall, our work establishes a foundation for making remotely hosted 3DGS content more quickly accessible to end-users in over-the-top consumption scenarios, with our results showing significant improvements in quality across all metrics compared to existing methods.
\end{abstract}
\begin{figure*}[t!]
    \centering
    \includegraphics[width=1.0\linewidth]{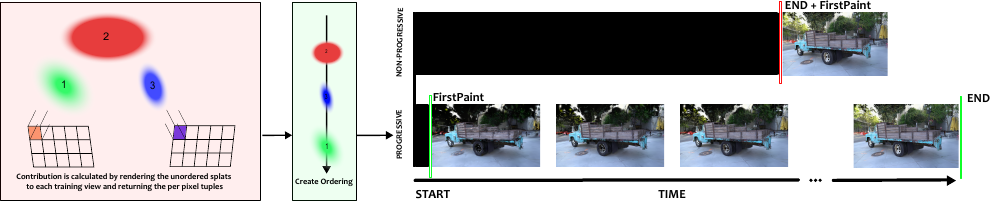}
    \caption{Our approach works by rendering all training views and returning the top 20 contributing splats per pixel per image. This information is used to create an ordering which can then be used to dynamically select how many splats are used. For progressive rendering/streaming, we can change the number of splats per sent chunk to start rendering as soon as possible without the user needing to wait. With little overhead in total time, we drastically reduce both time and bandwidth until time to First Paint.}
    \label{fig:approach}
\end{figure*}

\section{Introduction}
\label{sec:intro}
In recent years, radiance fields have attracted significant attention due to their ability to accurately portray real-life scenes while requiring only a limited number of input images. Neural Radiance Fields~\cite{mildenhall2020nerf} (NeRFs) played an important role in this resurgence as they allowed scenes to be represented implicitly by using a multilayer perceptron (MLP) to predict the volume density and view-dependent emitted radiance based on an input position and look-at direction. After the introduction of NeRF, follow-up work has focused on improving different aspects of the approach ranging from training time~\cite{mueller2022instant} and editing~\cite{liu2021editing} to other downstream tasks such as meshing~\cite{rakotosaona2023nerfmeshing}. Most of these approaches follow the same principles as NeRF and thus fall under the category of implicit approaches. Generally, they require less storage than explicit approaches at the expense of making downstream tasks less intuitive and often more difficult than explicit approaches, as there is a lack of direct manipulability. 

3D Gaussian Splatting (3DGS)~\cite{kerbl3Dgaussians} takes a different approach by shifting from an implicit to an explicit representation using 3D Gaussians. By optimizing a set of 3D Gaussians to represent a scene, 3DGS allows both fast training and inference, which has been pivotal in allowing a more widespread adoption of these techniques. Another important factor is that the format used to store these scenes is an extension of a regularly used point cloud format. This factor plays an important role in facilitating downstream tasks as it simplifies the process of experimenting with existing techniques as inspiration to bring them to the context of 3DGS. Our work will cover two downstream examples, \textit{progressive rendering} and \textit{progressive streaming}. Historically, progressive rendering has played an important role in content visualization, as visualizing the required content to the user as soon as possible is crucial in achieving a proper user experience. Progressive streaming focuses predominantly on the network distribution of media content, such as audio and video, where it is crucial that the user is able to consume the content as soon as possible. Despite this seemingly perfect fit, transferring existing progressive rendering methods to 3DGS is not trivial, as these do not fully utilize the information provided by 3DGS. Besides position, we additionally have access to the opacity, covariance, and view-dependent color information, which can all be used to achieve better results.

In our work, we thus bring progressive rendering (and, by extension, streaming) to the context of 3D Gaussian Splatting by utilizing a contribution-based prioritization method to determine a rendering order in post-training. We then integrate this with other contextual information, such as the current viewport, to ensure that relevant content is visualized to the user promptly. Concretely, using our approach allows us to reduce rendering delay and achieve faster time-to-first-paint times, which is expected to improve user experience. An overview of our approach is given in Figure~\ref{fig:approach}. By combining different commonly used graphics techniques together with our contribution-based ordering, we facilitate the visualization of the approximated scene using only a fraction of the data required to store the complete scene. We then incrementally use the remaining data to continuously improve quality over time (\cref{initfigure}). Our work is the first to bring progressive rendering of 3DGS into an academic context. We show that our proposed method outperforms available web-based approaches regarding quality per percentage splats used and show possibilities for further reductions in bandwidth utilization, using existing compression methods to bring progressive rendering to 3DGS scenes. Combining both compression and progressive rendering allows us to send an initial representative scene using only fraction storage required to store the complete uncompressed scene. Furthermore, our approach works for both complete scenes and on individual objects within a scene, paving the way for further application-specific research and applications.

In~\cref{sec:related_work}, we will discuss a selection of relevant works concerning progressive streaming, rendering, and compression of 3DGS scenes and sketch where our approach is situated within the domain. \cref{sec:method} will discuss the different aspects of our approach and how they contribute to the final result. \cref{sec:results} illustrates our results and performs a comparative study with existing web viewer-based approaches currently used within the community, an ablation study, the results for per-object ordering, and compression. Lastly,~\cref{sec:conclusion} outlines potential future trajectories and summarizes our key conclusions. 

Concretely, our contributions can be summarized as follows:
\begin{itemize}
    \item We propose an importance-based sampling method to decide the order of transmission that works on both the full scene and individual objects within a scene and is viewpoint-independent.
    \item We show an integration of progressive rendering into an existing compression method and show the potential for further bandwidth savings by using a combination of both.
    \item We show that our approach can be extended with frustum culling and an octree-based prioritization method to boost our performance further.
\end{itemize}

\section{Related Work}
\label{sec:related_work}
This section will briefly cover the related works of our approach. We will first discuss progressive rendering and streaming as they form the core contribution of our approach. We will then highlight several compression papers as both fields focus on quickly bringing content to the user. While compression achieves this by lowering the memory footprint, progressive approaches focus on sending the relevant content first.
\\

\textbf{Progressive Rendering of 3D content}
\quad
Progressive rendering of scenes has been an interest of research for decades as it plays a crucial role in facilitating the efficient rendering of 3D scenes. The content used has increased in both size and amount of detail over the years, necessitating more efficient usage of bandwidth and memory. Progressive rendering allows applications to start rendering a scene without the need for it to be fully loaded into memory or retrieved from a remote server first. Hoppe~\cite{10.1145/3596711.3596725} first introduced the term \textit{progressive} in 1996, specifically in the context of triangle meshes. He proposed to iteratively simplify a mesh using the edge collapse operation, which unifies two adjacent vertices into a single vertex. The simplest version is sent first, after which the inverse operation, vertex split, can be applied to fully reconstruct the original mesh. These inverse operations can be applied incrementally, allowing the application to progressively load the mesh. At a later stage, Hoppe~\cite{10.1145/258734.258843} revisited his earlier work and integrated view-dependent aspects into it. This updated approach used the viewing frustum, surface orientation, and screen-space projected error to take more context into account while reconstructing the mesh. More recently, Chen et al.~\cite{chen2023neural} have explored the use of neural networks to learn a progressively compressed representation for meshes. Their approach converts a given mesh into a datastream that can be progressively transmitted. The client can then decode this datastream to reconstruct a simplified mesh, which can be iteratively improved upon using subsequent transmissions. This process can continue until the original mesh is reconstructed or the required quality is reached.

Over the years, progressive rendering has been used for other representations besides meshes. Schütz et al.~\cite{pcprog} focus on progressively rendering point clouds. They achieve this by projecting the previous set of points onto the viewing frustum and only transmitting points that appear in unoccluded areas of the new frame. In the context of radiance fields, BungeeNeRF~\cite{xiangli2022bungeenerf} allows progressive rendering of NeRFs for multi-scale scene rendering. The network progressively scales with the scale of the learned scene, aiming to divide quality into different layers. Reconstruction quality can then be decided by querying a different head that corresponds to a certain quality. 


\textbf{Progressive Streaming of 3D content}
\quad
The distinctive difference between progressive rendering and progressive streaming is that the latter also involves the network delivery of the media content from a remote server to the consuming client device. As such, with progressive streaming, the client progressively renders the content at the rate it receives from the content server. This concept is widely adopted in the over-the-top delivery of conventional audiovisual content via the HTTP Adaptive Streaming (HAS) paradigm and its standardized MPEG-DASH implementation~\cite{isoiec22mpegdash, sodagar11mpegdash}. Recently, the use of progressive streaming is also being explored for more immersive media formats. Noteworthy examples are the HAS-like progressive streaming of textured geometry (e.g., \cite{lemoine23gltfStreaming, slocum21via, farrugia23framework, lievens21RelevanceABR, forgione18dash3D, zampoglou16adaptive_web3d}), light fields (both static~\cite{wijnants18HASSLF, lievens21SLFInWeb} and dynamic~\cite{hu23lfvideoGNN, kara18evaluate_dynamic_adaptive_lf}) and point clouds (e.g., via the MPEG V-PCC specification\cite{isoiec23vpcc, vanderhooft19has_pcc}). With respect to the progressive streaming of radiance fields, the attention of the academic community has focused mostly on the streaming of NeRFs representing scenes that are static (e.g., Cho et al.\cite{cho22streamableNeRF}). Of specific interest is the NeRFHub approach by Chen et al. \cite{chen24NeRFHub}, which aims to minimize the network transfer latency of NeRF models while satisfying lower limits regarding rendering smoothness and perceptual quality. NeRFHub does so by (amongst others) shrinking the number of hidden MLP layer channels (in combination with selective model training) and by quantizing the feature grid's floating point values. While NeRFHub shares our objectives of reducing start-up latency during over-the-top radiance field consumption, no progressive streaming nor rendering is applied, as NeRFHub always deals with (quality-variant) integral NeRF models. Finally, to date, progressive 3DGS streaming remains largely unexplored in academic literature; in effect, this paper represents a fundamental step in that direction.

\textbf{Compression of 3D Gaussian Splatting}
\quad
Compression and progressive streaming share a similar goal of reducing the bandwidth needed to render a scene. While progressive streaming focuses on sending relevant content first, the goal of compression is to minimize the total memory footprint. As this goal is very similar, and both approaches can be used in conjunction with each other, we will briefly overview the relevant works in this field. 

We argue that there are two main ways of performing compression of 3DGS. The first way focuses on compressing a scene while maintaining all existing parameters. Codebooks and vector quantization are often used in these approaches as they allow existing data to be stored with less memory, albeit with some loss in accuracy. The second way focuses on adapting the actual parameters themselves to a more efficient format. Niedermayr et al.~\cite{niedermayr2023compressed} performed compression by utilizing a codebook to store Gaussian parameters. After converting an existing scene into its codebook representation, they fine-tune all parameters using the training images to regain lost quality due to discretization. Further compression is achieved by ordering the Gaussians and using run-length encoding to create the final representation. Compact3D~\cite{navaneet2023compact3d} works under the assumption that large groups of Gaussians will be likely to share parameters. They then use K-means clustering to group similar Gaussians and use a codebook to store data per cluster. They also promote fewer Gaussians in a scene by adjusting the opacity values to be close to one or zero. EAGLES~\cite{girish2024eaglesefficientaccelerated3d} uses a technique similar to our own as they also use contribution as a metric to filter splats. Besides this, they use vector quantization to reduce storage further. The previously mentioned approaches all fall under the first category and focus on compressing the existing parameters into a smaller format. Papantokankis et al.~\cite{papantonakis2024rmfsplat} propose different methods in their work, such as resolution-aware pruning and adaptively adjusting the number of coefficients used to model directional radiance. LightGaussian~\cite{fan2023lightgaussian} also focuses on the directional radiance and distills spherical harmonics to a lower degree, thereby compressing the largest contributor to memory usage. Our approach can be used together with any compression technique that allows for some factorization based on the contribution of a Gaussian to a given pixel.
\section{Methododology}
\label{sec:method}
Our approach allows progressive rendering of 3DGS by determining an ordering among Gaussians. By prioritizing Gaussians that have contributed significantly to the scene, we reduce the number of bytes needed for a qualitative approximate reconstruction. By rendering each training viewpoint and calculating the contribution across all views, we create an initial ordering of the Gaussians. We further refine this by inserting all Gaussians into an octree and selecting the top contributing Gaussians per leaf node. This results in the splats being more evenly spread across the scene, with a small hit to the quality of the foreground. Additionally, we can enable frustum culling to further increase the perceptual quality of the visible scene. Besides creating an order at the scene level, our approach also functions at an object level. This allows us to prioritize splats within individual objects or to prioritize objects within the scene. Our final contribution is the integration of our technique into an existing compression method to fully demonstrate the capabilities of this combination when it comes to reducing storage and bandwidth requirements. In the following subsections, we will discuss each aspect of our approach in more detail.

\subsection{Contribution-based ordering}
\label{subsec:formula}
The fundamental idea behind our approach comes from using the contribution of a Gaussian to all rendered views as a metric of its importance to the scene. Large Gaussians with a high opacity value will most likely contribute more than smaller splats with a low opacity. Using this intuition, we render all training views and calculate the contribution of a Gaussian across them to use as a contribution score. Sorting based on this contribution score leads to an ordering among Gaussians that estimates how important each Gaussian is for reconstructing the final scene. To retrieve the actual contribution value, we render the image and store the weighting factor of the top K contributing Gaussians per pixel. The color C of a pixel is determined as:
\[C = \sum_{i=1}^{N}T_i\alpha_ic_i\]
with
\[ T_i=\prod_{j=1}^{i-1}(1-\alpha_i)\] 
where $T_{i}$ represents the transmittance, $\alpha_i$ the opacity value, $c_i$ the color value, and N loops over the Gaussians overlapping the current pixel. The contribution is then taken as $T_i\alpha_i$.
For each pixel, we return a pairing of contribution and Gaussian ID to measure the final contribution of a specific Gaussian as:
\[B_{id}=\sum_{v\in\mathcal{V}}\sum_{p\in\mathcal{P}}(T_{id}\alpha_{id})\]
where $\mathcal{V}$ represents the set of viewpoints and $\mathcal{P}$ represents all pixels per viewpoint. This contribution directly translates into an ordering that reflects the importance of a Gaussian to the scene. It can be used to create chunks of Gaussians that can be rendered or sent independently of each other. EAGLES~\cite{girish2024eaglesefficientaccelerated3d} uses the same metric during training to prune splats that are less impactful during training. Instead, our approach opts to use it in post-training as a way to select impactful splats that are advantageous to send first.
\begin{figure*}
    \centering
     \includegraphics[width=1.0\linewidth]{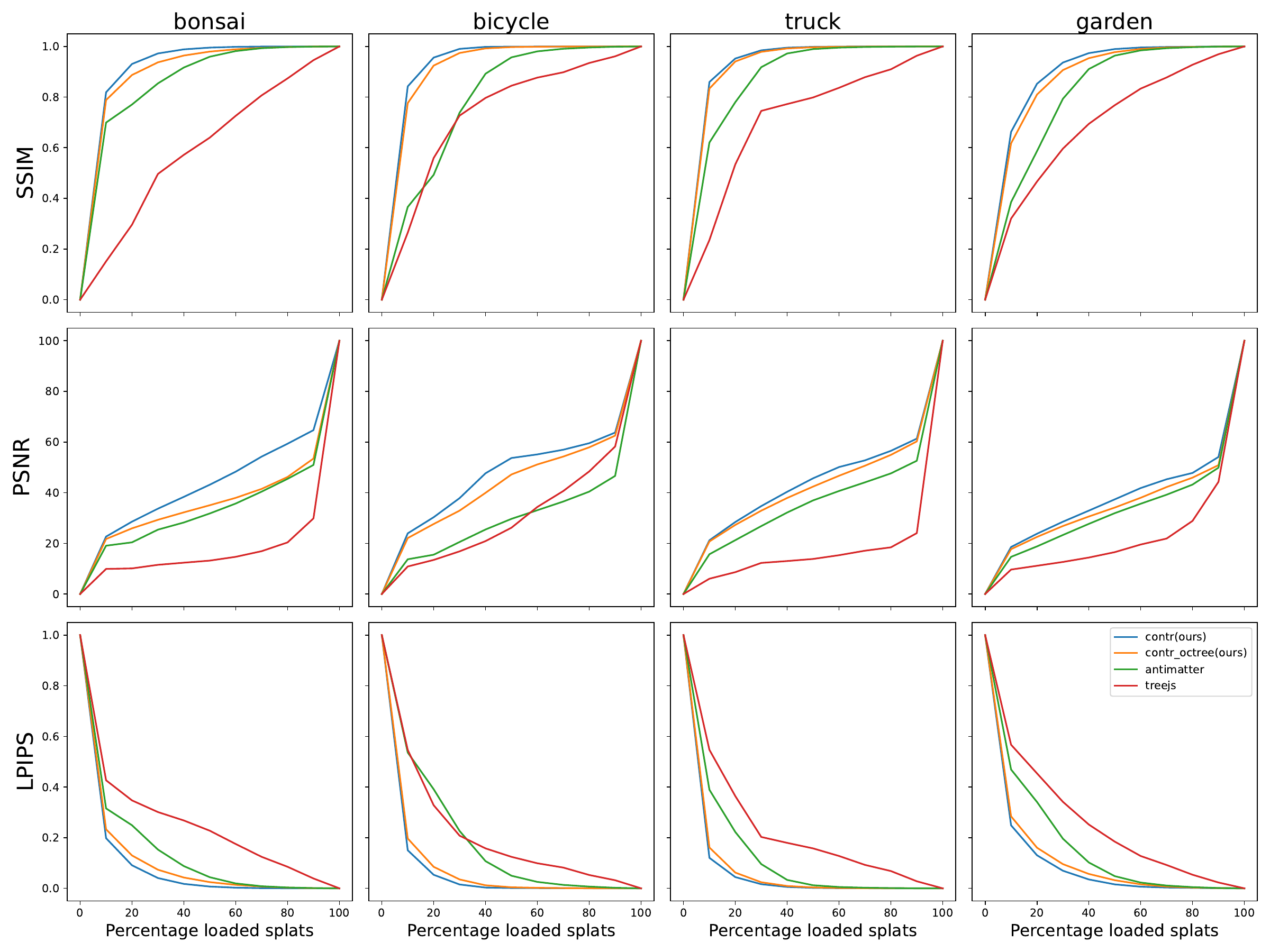}
    \caption{We show the results of different scenes for all three metrics and that our two approaches consistently outperform the previous methods by a significant margin. These scenes are taken from the MipNeRF360~\cite{barron2022mipnerf360} indoor and outdoor datasets and the Tanks\&Temples~\cite{Knapitsch2017} dataset. PSNR is measured compared to the original reconstruction and is capped at 100 when the complete scene is loaded in. (PSNR$\uparrow$, SSIM$\uparrow$, LPIPS$\downarrow$)}
    \label{fig:metrics_comp_packed}
\end{figure*}
\subsection{Refinement of ordering using octree}
\label{subsec:octree}
Using the global ordering created by the initial step, the chosen splats tend to focus on areas densely captured in the input data (see \cref{sec:results}). As we sum up all pixels, Gaussians that have been seen more often will be more likely to have a higher contribution score. This can result in sparsely captured parts being unfairly under-reconstructed. To counter this effect, we utilize an octree to create a spatial subdivision in the scene based on the density of splats. By taking the highest contributing splats per leaf node, we effectively spread out the Gaussians over the scene and thus make sure every part is partly reconstructed. Changing the maximum depth of the octree changes how much denser areas are prioritized. Density, in most cases, corresponds to areas of higher detail. This, however, does not always correspond perfectly to the distinction between background and foreground; thus, caution is needed when setting this depth to not over or underprioritize these parts of a scene. In \cref{sec:results}, we show the impact of this parameter in more detail.
\begin{figure*}[b]
    \centering
    \begin{subfigure}[b]{0.24\linewidth}
        \centering
        \includegraphics[width=\linewidth]{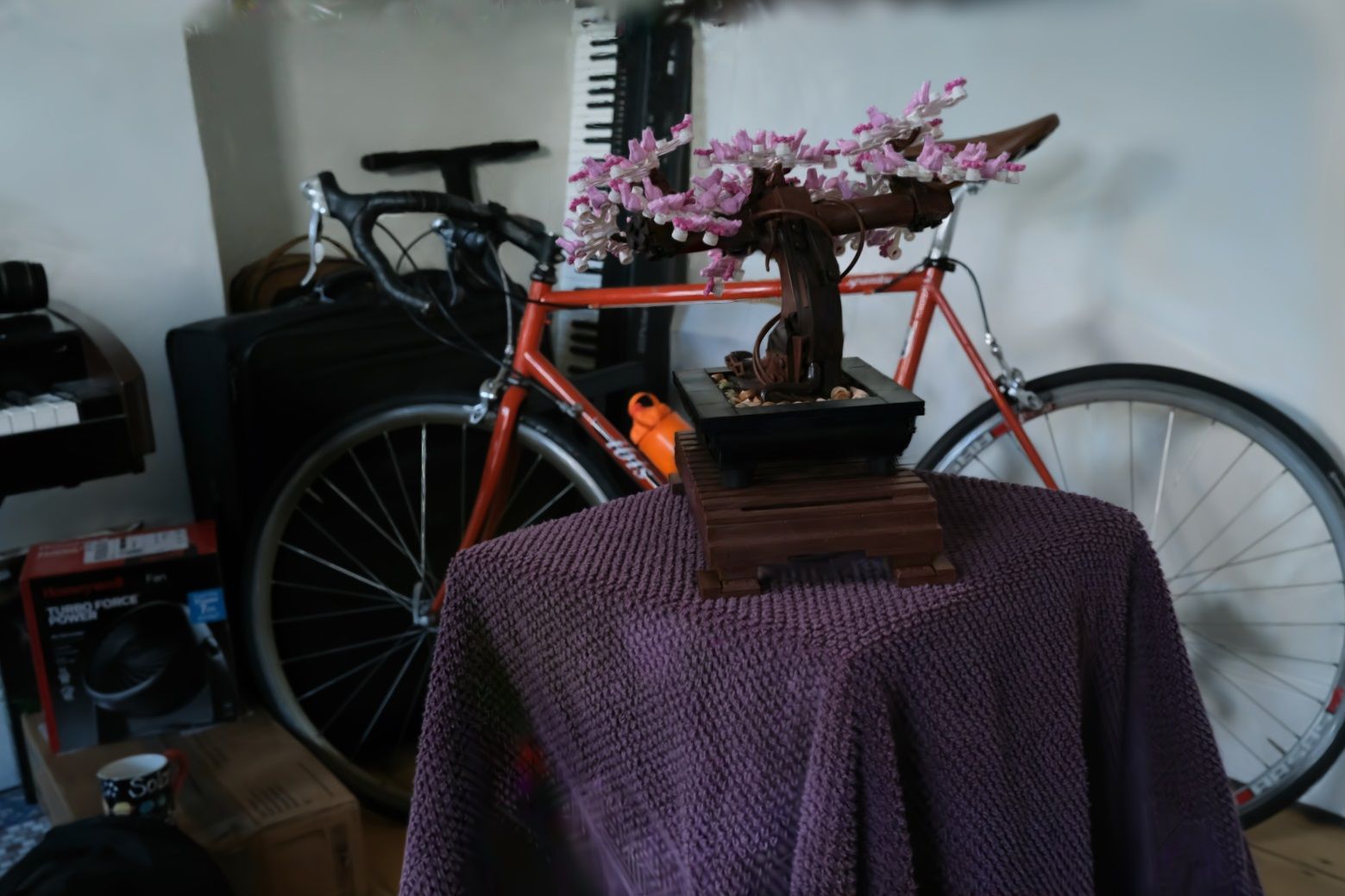}
        \caption{Ground Truth}
        \label{fig:gt_bonsai}
    \end{subfigure}
    \hfill
    \begin{subfigure}[b]{0.24\linewidth}
        \centering
        \includegraphics[width=\linewidth]{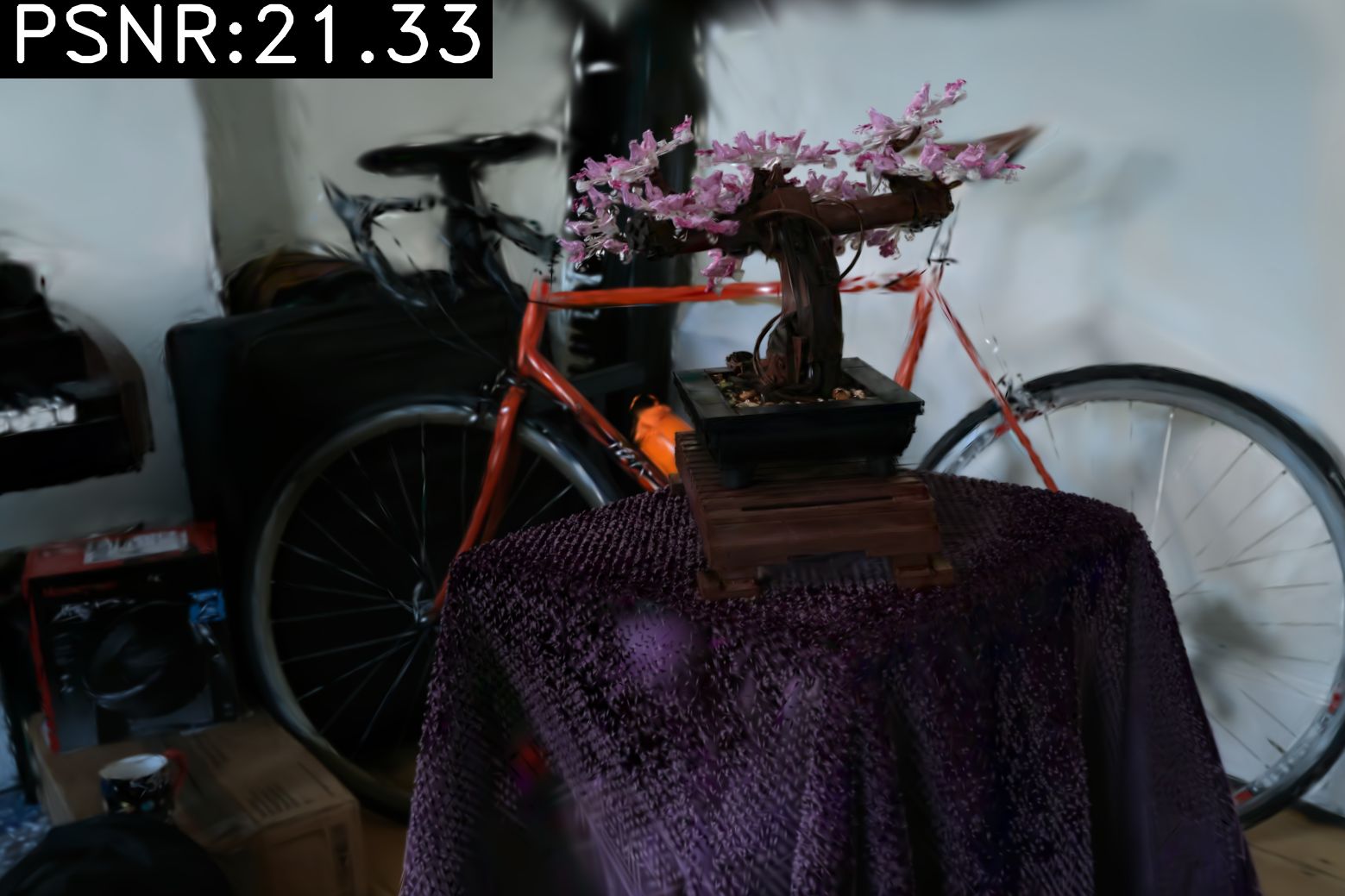}
        \caption{Ours}
        \label{fig:ours_bonsai}
    \end{subfigure}
    \hfill
    \begin{subfigure}[b]{0.24\linewidth}
        \centering
        \includegraphics[width=\linewidth]{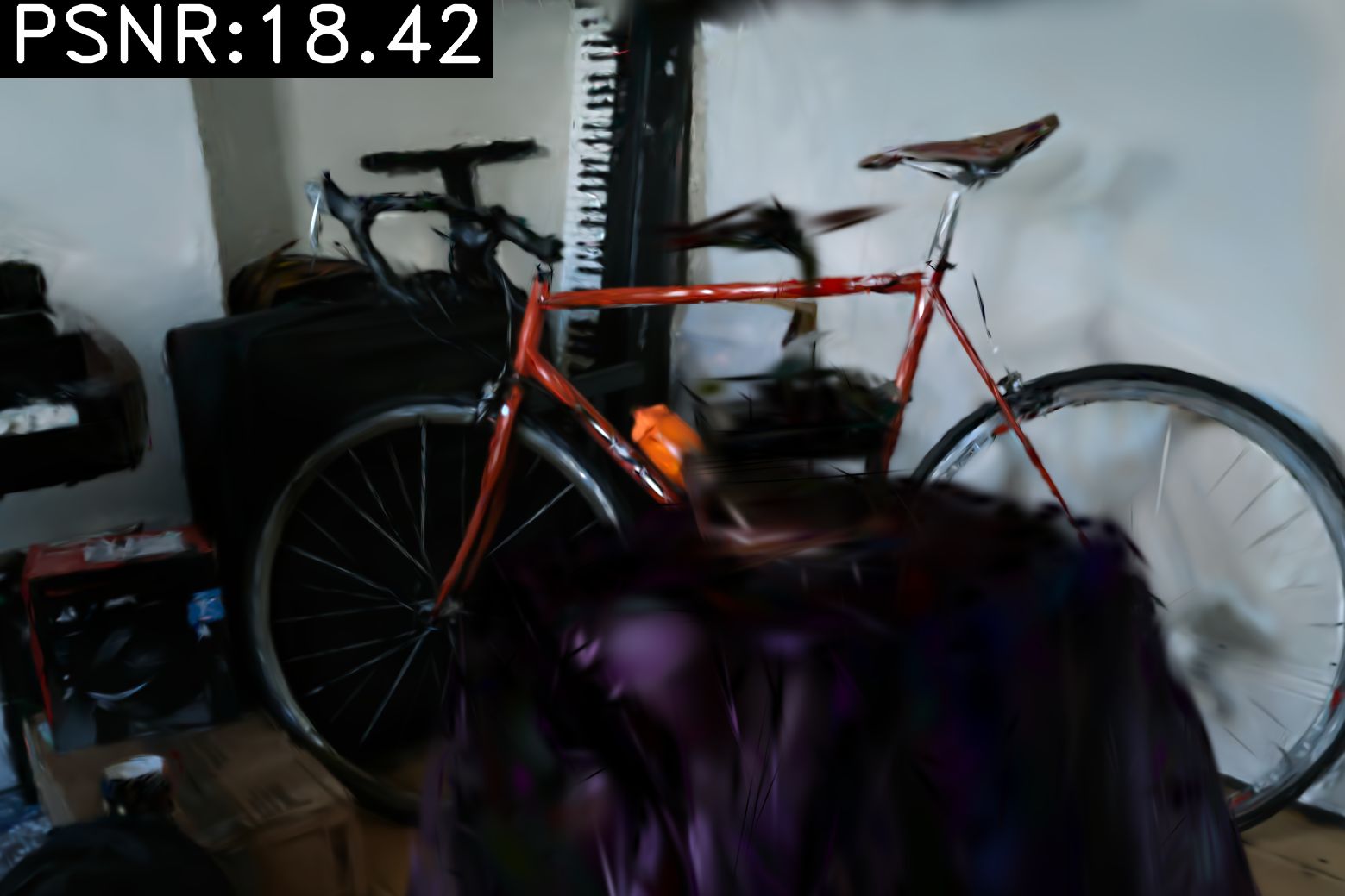}
        \caption{Antimatter}
        \label{fig:antimatter_bonsai}
    \end{subfigure}
    \hfill
    \begin{subfigure}[b]{0.24\linewidth}
        \centering
        \includegraphics[width=\linewidth]{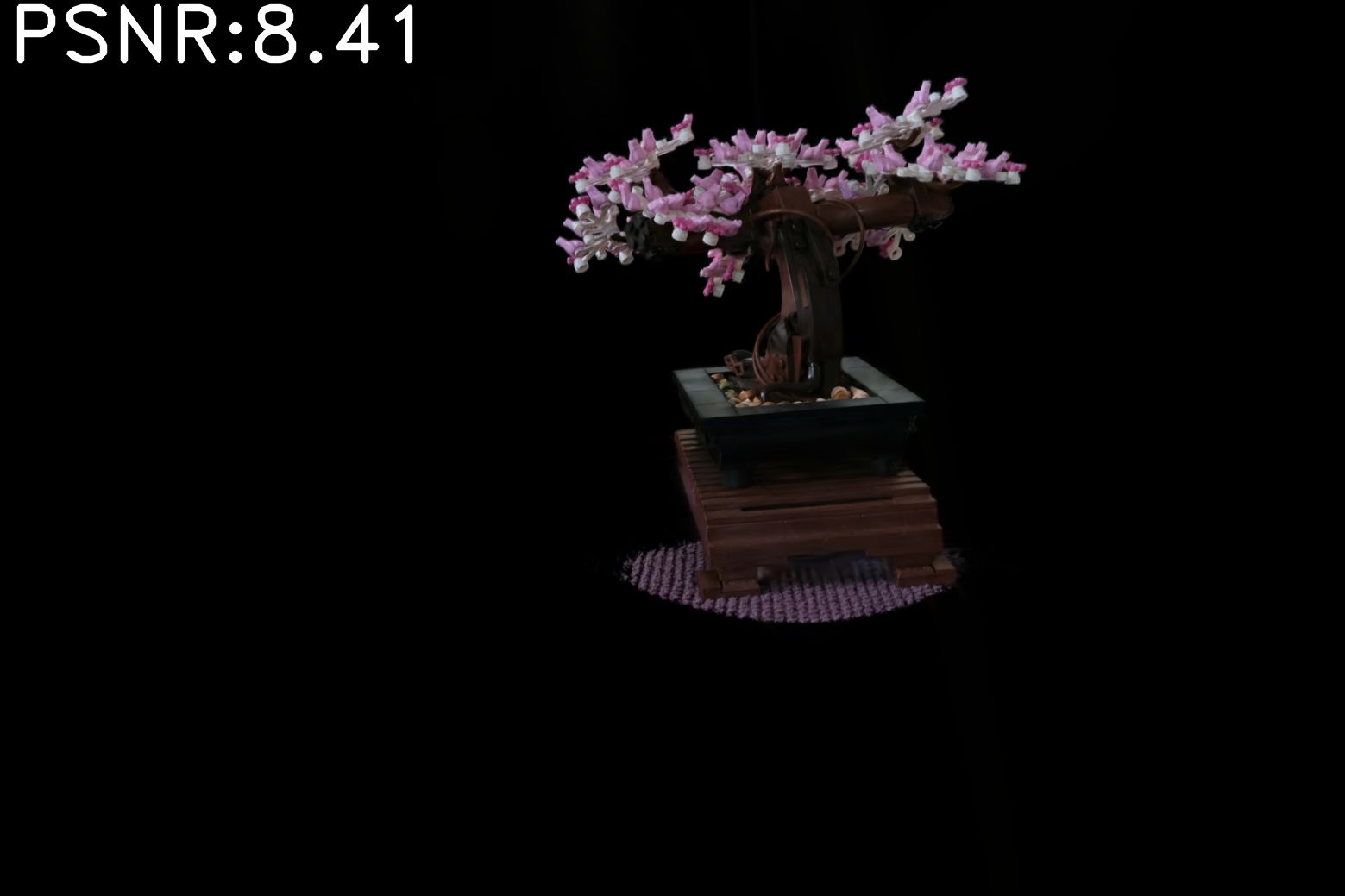}
        \caption{Three.js}
        \label{fig:treejs_bonsai}
    \end{subfigure}

    \caption{A concrete comparison between the three aforementioned methods on the \textit{bonsai} scene taken from Mip-NeRF360 indoor scenes being rendered at 10\% splats compared to the ground truth. Notice how both the bonsai and cloth are being rendered at a higher quality in Ours(b) than in Antimatter(c). Tree.js(d) will almost always be outperformed using standard metrics as the background is completely missing.}
    \label{fig:metrics_comp}
\end{figure*}
\subsection{In-frustum prioritization}
As the initial pose of the user is known beforehand, we can use it to fine-tune our initial ordering further. Using our initial list of Gaussians, we can select a percentage of splats chosen from within this frustum to allow rapid movement while focusing mainly on what is visible to the user. To determine whether or not a splat lies inside the viewing frustum, we project all splats onto the image plane and filter the splats that fall outside it. As Gaussians cover areas and not points on the image plane, there is a chance that a Gaussian of which the mean falls outside the frustum still contributes to the image. We thus follow 3DGS and allow a small margin around the frustum in which we still count Gaussians as being inside. 

\subsection{Object-level rendering}
By segmenting splats that belong to a specific object, we can create an ordering within an object or compare objects at a scene level based on the sum of their contributions. Within an object, we apply the same principles as at the scene level and prioritize splats that have a significant contribution to the object. Within a given scene, we can prioritize objects based on their contribution to the scene and send higher-quality representations of important objects while sending lower-quality, low-contributing splats. In the context of progressive rendering, both these cases are important to have as they allow more advanced progressive rendering/streaming techniques to be implemented. For now, we achieve this by manually segmenting objects or using existing 2D segmentation methods~\cite{ravi2024sam2} and finding the contribution of splats within this segmentation to determine the importance of objects for a given scene. This can, however, also be used for scenes that are composed of different trained models.

\subsection{Integration into compression methods}
As our approach is agnostic to any Gaussian-specific parameter and can use solely contribution to calculate an ordering, our approach can be applied to other 3DGS-inspired approaches. We have to point out that our approach only has benefits when per-Gaussian data is stored, which despite the frequent use of codebooks, is the case for the majority of compression papers. To show this claim holds up, we integrate our approach into the approach of Niedermayr et al.\cite{niedermayr2023compressed} and demonstrate that our approach achieves similar orders of progressive rendering gains. We simply adapted the renderer to include passing back the contribution values and IDs per Gaussian and calculating our ordering afterward. Caution has to be taken as this approach sorts the Gaussians in Morton order before saving them, which completely voids our previous ordering. Furthermore, when vector quantization is done using the minimum and maximum values, they have to be passed as well to allow for dequantization.

\section{Experiments}
\label{sec:results}
\textit{Dataset}. We used the MiPNeRF360~\cite{barron2022mipnerf360} dataset (inside \& outside), two scenes from the Tanks\&Temples\cite{Knapitsch2017} dataset, and two scenes from the Deep Blending~\cite{DeepBlending2018} dataset to validate our approach. We used the original code of 3DGS~\cite{kerbl3Dgaussians} to split the data into train and test sets to promote consistency.

\textit{Evaluation metrics}. We use community-standard image loss metrics Peak-signal-to-noise ratio (PSNR), Learned Perceptual Image Patch Similarity (LPIPS), and Structural Similarity Index (SIMM) to measure the visual quality. We compare each approach to the final trained approach and not the ground truth to show how well it approximates sending the complete trained scene.

\textit{Implementation details}. For our experiments, we used the contribution of the 20 highest contributing Gaussians per pixel to decide on the ordering, which is an overestimation in most cases. In experiments using an octree (\cref{subsec:octree}), we use a depth of 3 unless otherwise stated. For frustum culling, we allow some margin by increasing the frustum boundaries by 30\%. For experiments with specific objects within a scene, we manually segmented them from the trained scene. We used the approach by Niedermayr et al.~\cite{niedermayr2023compressed} for all experiments concerning compression and adapted their approach to allow for chunking of the compressed files into different independent files based on contribution to the training views. Concretely, we insert the minimum and maximum splats to allow for dequantization when loading the .ply files. 

\subsection{Quantitative Evaluation}
\label{subsec:base}
As our work is the first to bring progressive rendering of 3DGS into the academic context, we compare ourselves with widely used online web viewers. We also show several variants of our own work and verify that our approach consistently outperforms other approaches. Concretely, we compare our implementation to \textit{3D Gaussian Splatting with Three.js}\footnote{https://github.com/mkkellogg/GaussianSplats3D} and \textit{3D Gaussian Splatting viewer by antimatter15}\footnote{https://github.com/antimatter15/splat}. Most other available viewers are closed-source or require the complete .ply file to be downloaded before rendering starts, which can take from multiple seconds up to several minutes. This further demonstrates the need for open-source progressive rendering approaches. Antimatter's viewer prioritizes splats based on a combination of their opacity and scale:  
\begin{center}
    $G_{contr}=\frac{-e^{scale.x + scale.y + scale.z}}{1+e^{-opacity}}$
\end{center}
while the three.js implementation opts to load in splats starting from the center. We can simply use the Euclidean distance from the center as a metric of contribution:
\begin{center}
    $G_{contr} =\sqrt{(x_2 - x_1)^2 + (y_2 - y_1)^2 + (z_2 - z_1)^2} $
\end{center}
where the center is taken as (0,0,0). Our approach is calculated as in \cref{subsec:formula}.
In \cref{fig:metrics_comp_packed}, we show that our approach outperforms previous methods by a significant margin across different datasets and metrics. The images in \cref{fig:metrics_comp} also demonstrate a clear perceptual improvement of our approach. The \textit{bonsai} and \textit{tablecloth} are loaded in at a higher quality while using the same amount of splats. Large splats that contribute to multiple pixels will inherently get a higher score and, thus, are more likely to be sent first. As each pixel's contribution is capped at 1, smaller splats are inherently disadvantaged and will need a large contribution to receive a higher overall score. We argue that these attributes are wanted behavior which is proven by \cref{fig:metrics_comp} clearly showing a large jump in quality over the previous methods.
\begin{figure}[t]
    \centering
    \begin{subfigure}[t]{0.49\linewidth}
        \centering
        \includegraphics[width=\linewidth]{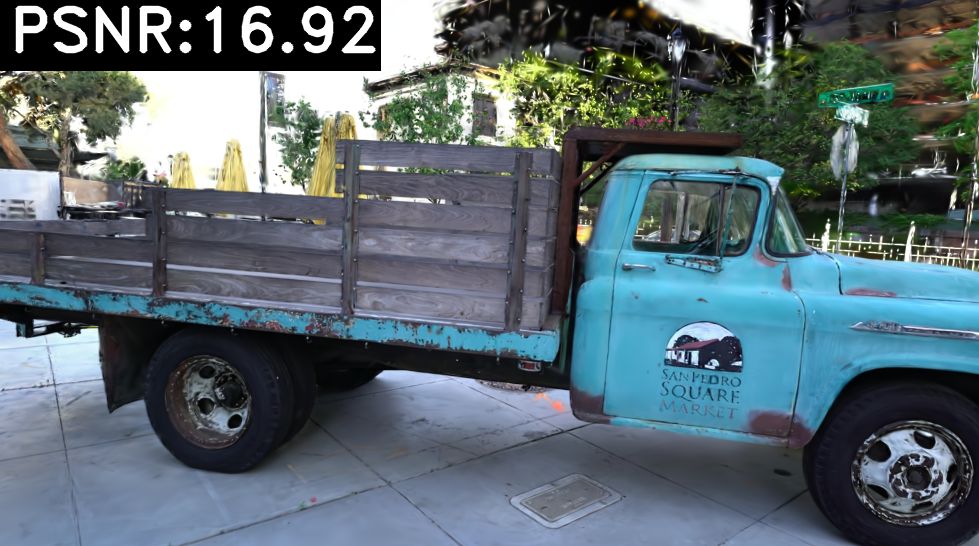}
        \caption{Base}
        \label{fig:train_contr_with_psnr}
    \end{subfigure}
    \hfill
    \begin{subfigure}[t]{0.49\linewidth}
        \centering
        \includegraphics[width=\linewidth]{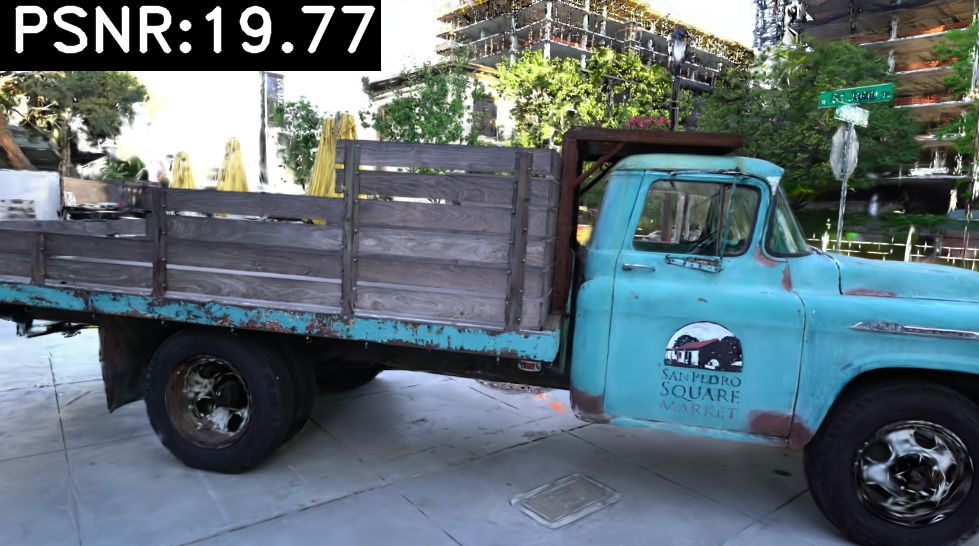}
        \caption{Base+Octree}
        \label{fig:train_contr_octree_with_psnr}
    \end{subfigure}

    \begin{subfigure}[t]{0.49\linewidth}
        \centering
        \includegraphics[width=\linewidth]{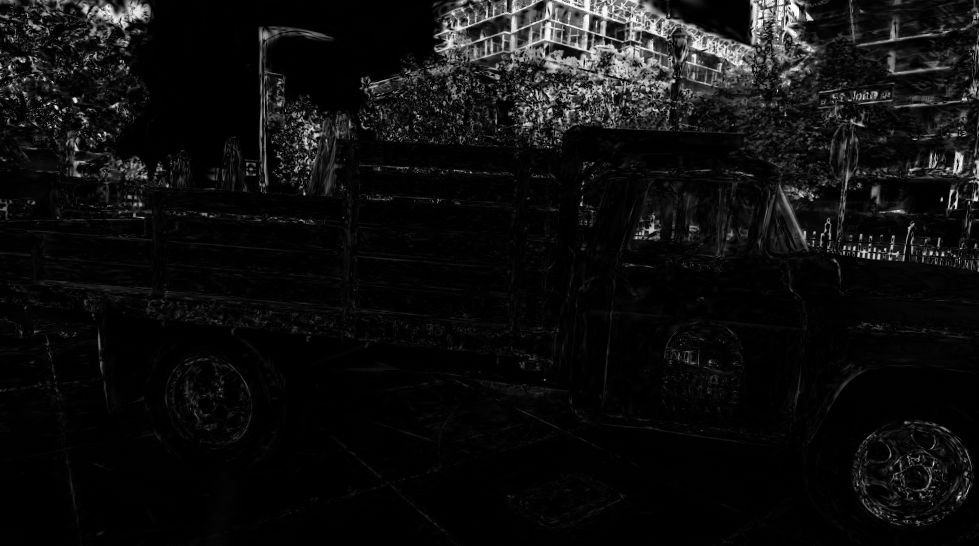}
        \caption{Difference Base \& GT}
        \label{fig:contr_diff}
    \end{subfigure}
    \hfill
    \begin{subfigure}[t]{0.49\linewidth}
        \centering
        \includegraphics[width=\linewidth]{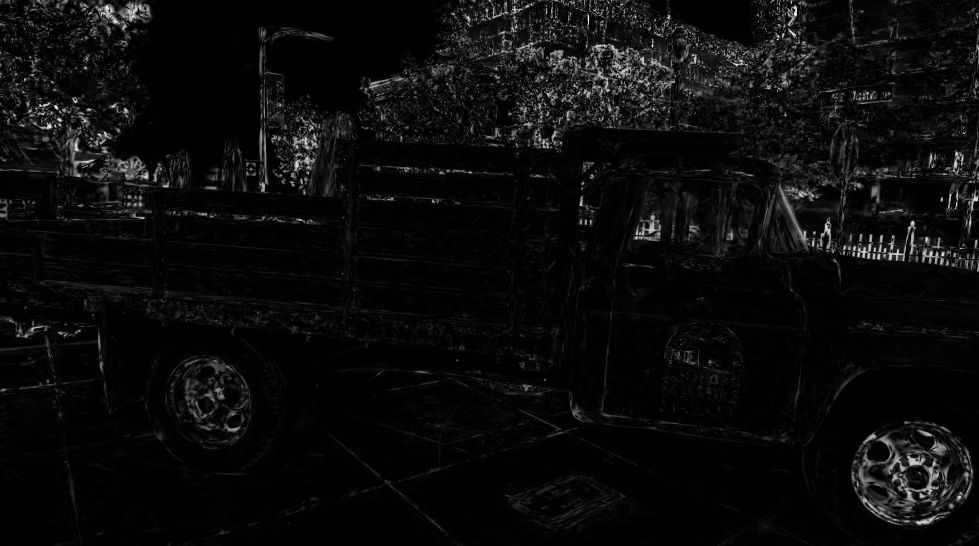}
        \caption{Difference Base+Octree \& GT}
        \label{fig:octree_diff}
    \end{subfigure}
    
    \caption{Despite the right image (b) performing better than the left image (a), we can clearly see there are more errors in the foreground when comparing (c) with (d). Due to most images containing mostly containing closer areas, the gains we achieve in the background are not reflected in the average PSNR scores for the complete scene, as was shown in \cref{fig:metrics_comp_packed}. Perceptually, our base approach with the octree performs better overall.}
    \label{fig:octree}
\end{figure}
\subsection{Ablation Study}
\begin{figure}[t]
    \centering
    \begin{subfigure}[b]{0.49\linewidth}
        \centering
        \includegraphics[width=\linewidth]{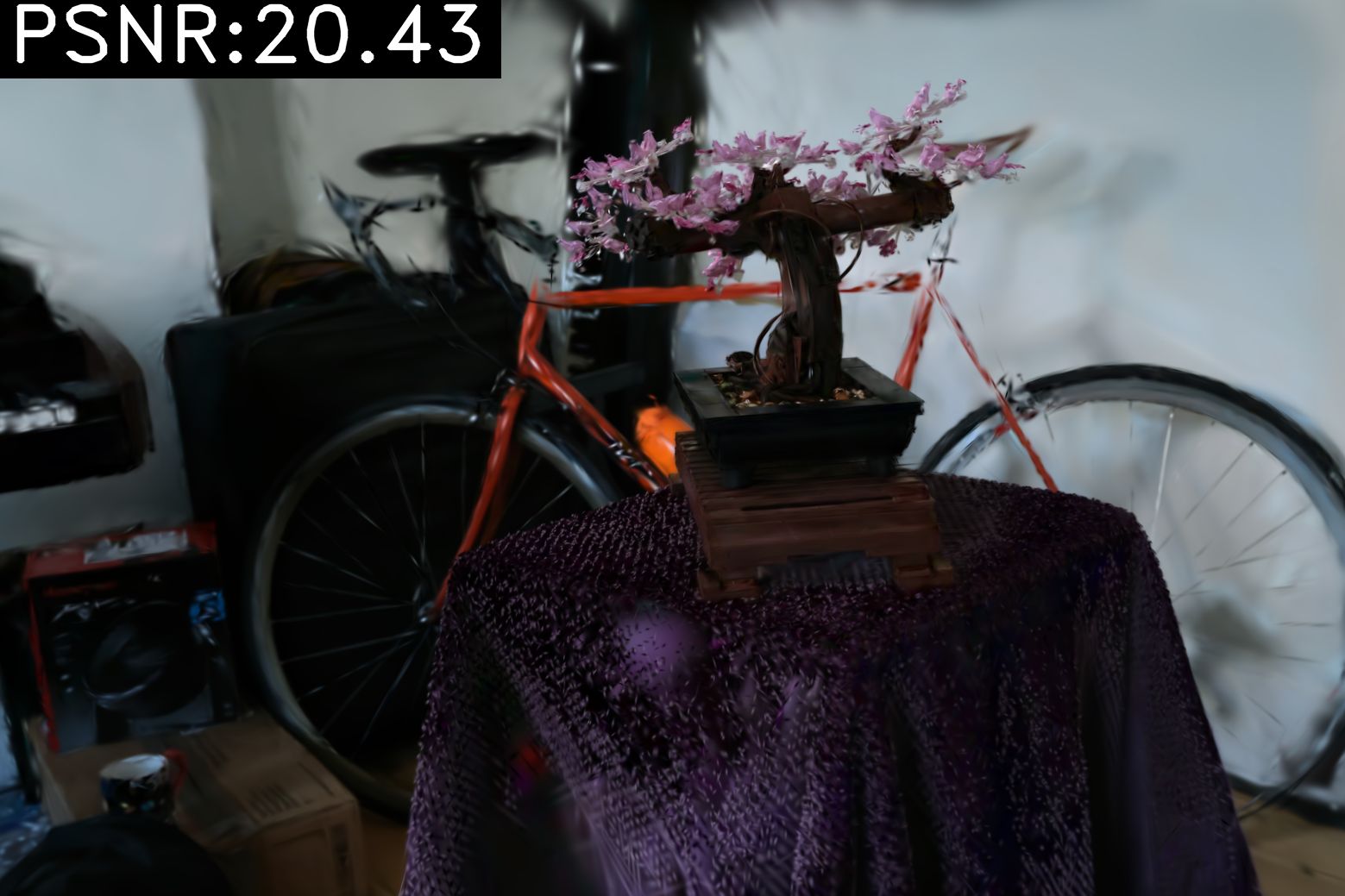}
        \caption{Depth 1}
        \label{fig:base_approach}
    \end{subfigure}
    \hfill
    \begin{subfigure}[b]{0.49\linewidth}
        \centering
        \includegraphics[width=\linewidth]{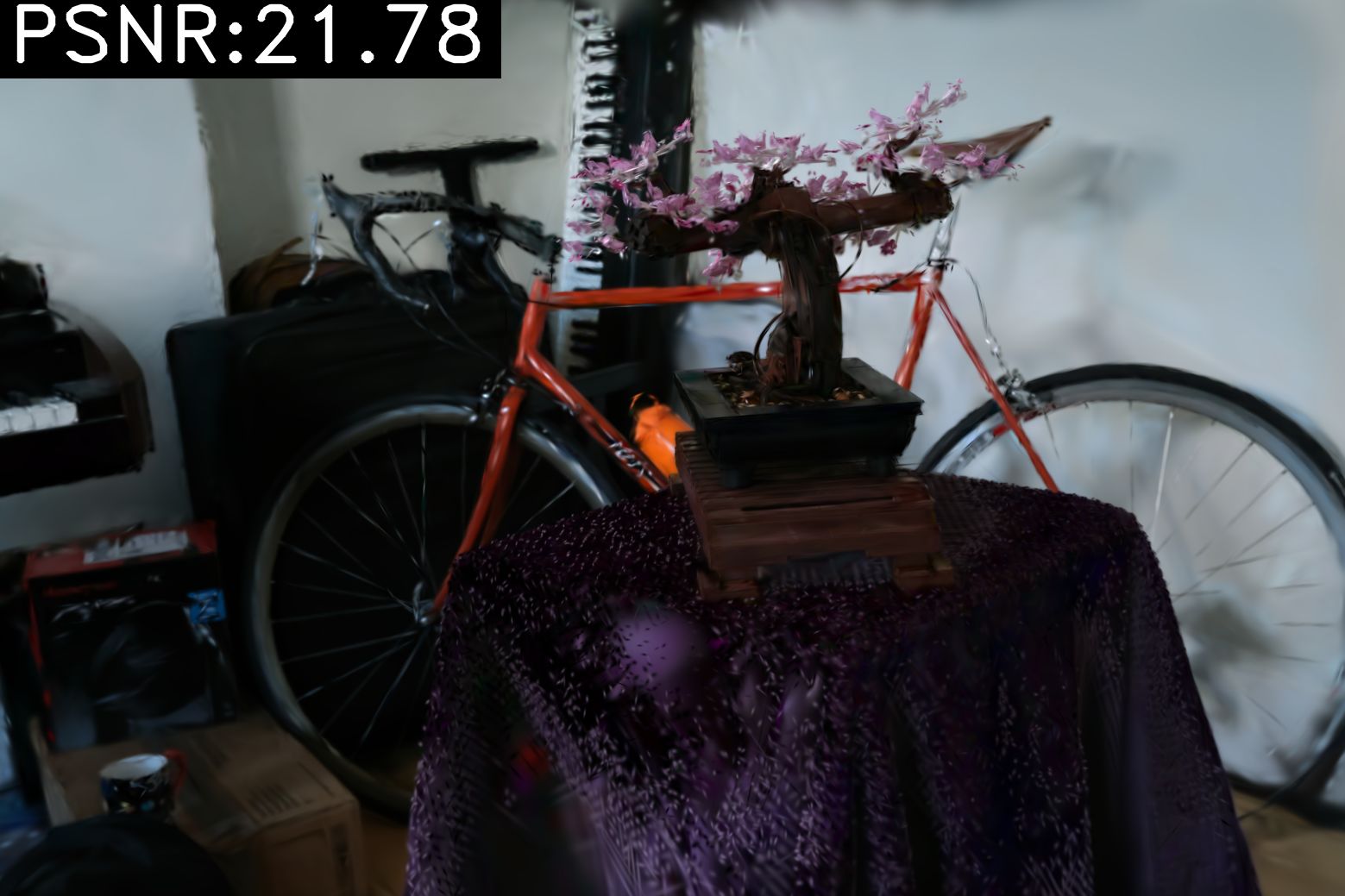}
        \caption{Depth 3}
        \label{fig:intermediate_approach}
    \end{subfigure}
    \vspace{0.5em}
    \begin{subfigure}[b]{0.49\linewidth}
        \centering
        \includegraphics[width=\linewidth]{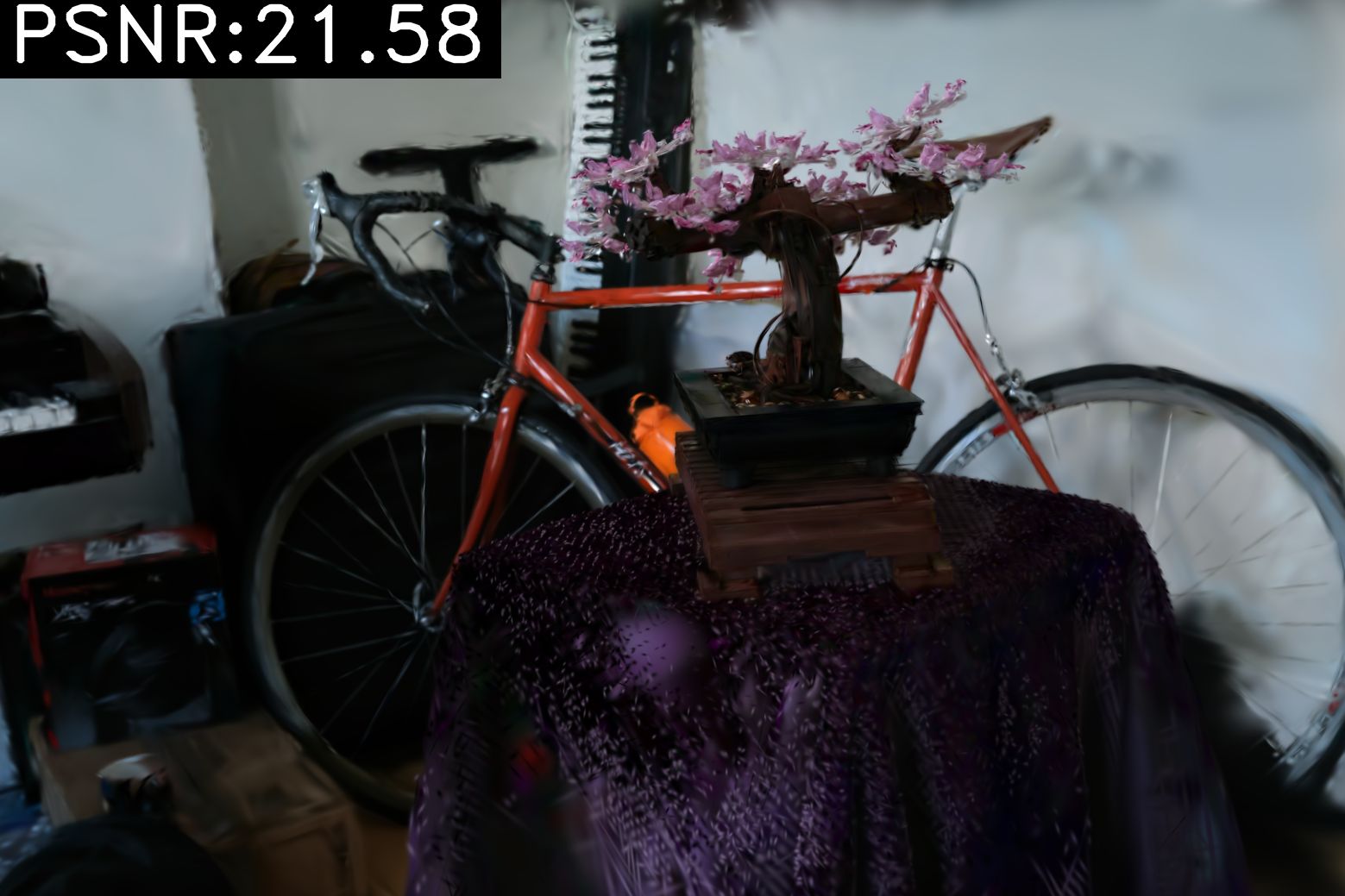}
        \caption{Depth 5}
        \label{fig:advanced_approach}
    \end{subfigure}
    \hfill
    \begin{subfigure}[b]{0.49\linewidth}
        \centering
        \includegraphics[width=\linewidth]{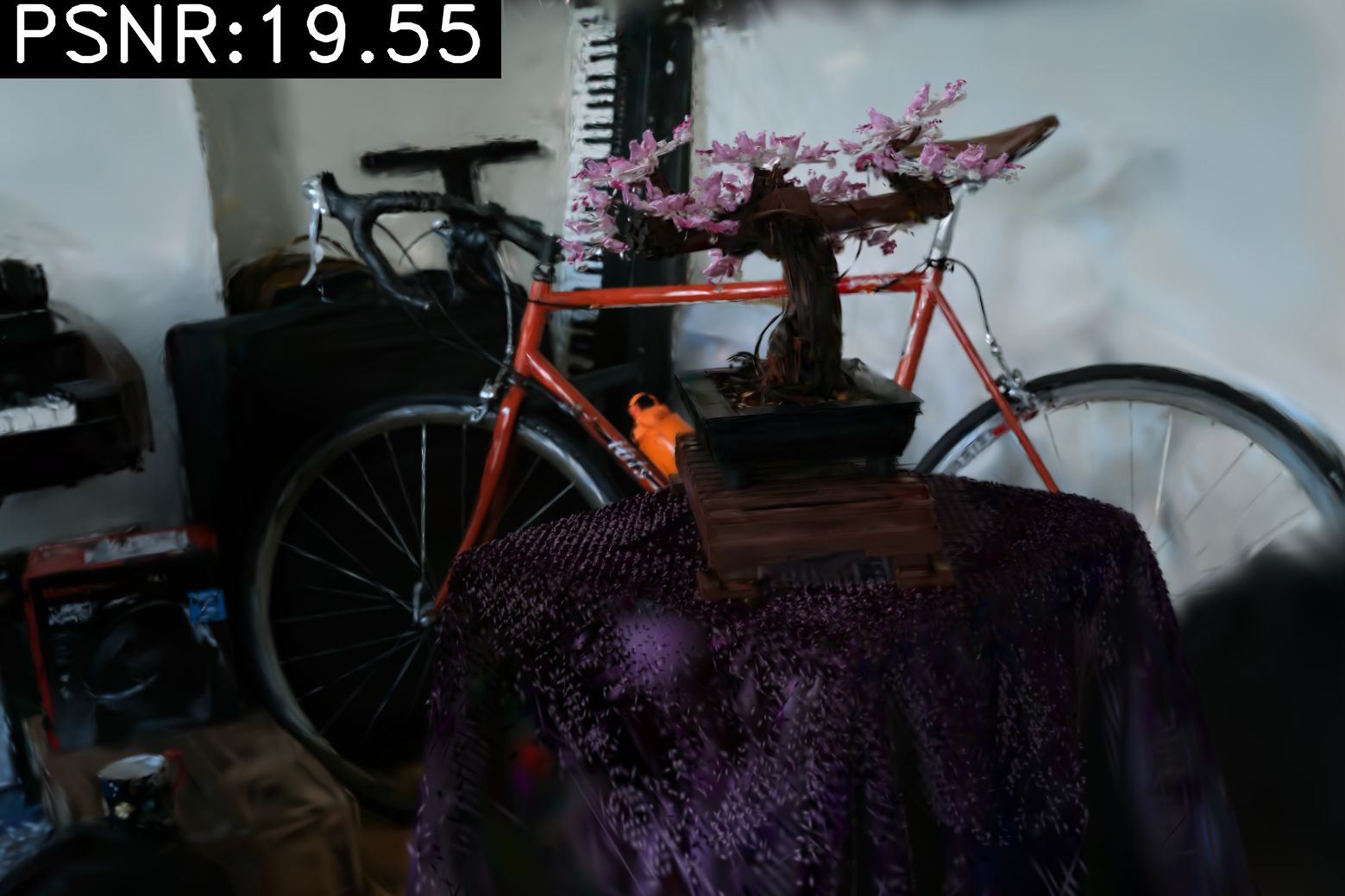}
        \caption{Depth 7}
        \label{fig:complete_approach}
    \end{subfigure}
    \caption{We show the impact of the octree depth parameter on the rendering quality. Using a depth that is either too low or too high will result in not enough attention being given to the background. Using 3 as the depth parameter consistently yields good results across all scenes.}
    \label{fig:octree_depth}
\end{figure}

The base implementation (see~\cref{subsec:base}), which solely used the contribution across pixels and views as a metric, achieves state-of-the-art results. We will now show the impact of the aforementioned additions to this base implementation and their impact on the final rendering quality. Our first addition is the octree, which forces Gaussians to spread out more over the scene. This results in the sparsely captured parts of the scene being loaded in sooner, as can be seen in \cref{fig:octree}. On average, including the octree approach results in a loss in PSNR as the foreground often contributes more to PSNR. Perceptually, however, we observe a clear difference between both images. Considering the difference images in \cref{fig:octree}, the background shows fewer errors while the foreground becomes slightly worse. This deterioration of the foreground results in lower PSNR scores on average. For now, we leave this as a subjective observation, and future user studies will be needed to verify our intuition that using the octree does, in fact, lead to a better user experience. \cref{fig:octree_depth} then shows the impact of the depth parameters, clearly illustrating the negative impact of selecting a depth that is too high. Lower depth values have little to no impact, as can be seen by keys on the keyboard not being present. Medium depth values, such as 3 and 5, perform well in both the foreground and the background. A value of 7 starts introducing errors, as can be seen by the black blur on the bottom right.
As this value gets larger, it comes closer to the base approach. 

\begin{figure}[t]
    \centering
    \begin{subfigure}[b]{0.49\linewidth}
        \centering
        \includegraphics[width=\linewidth]{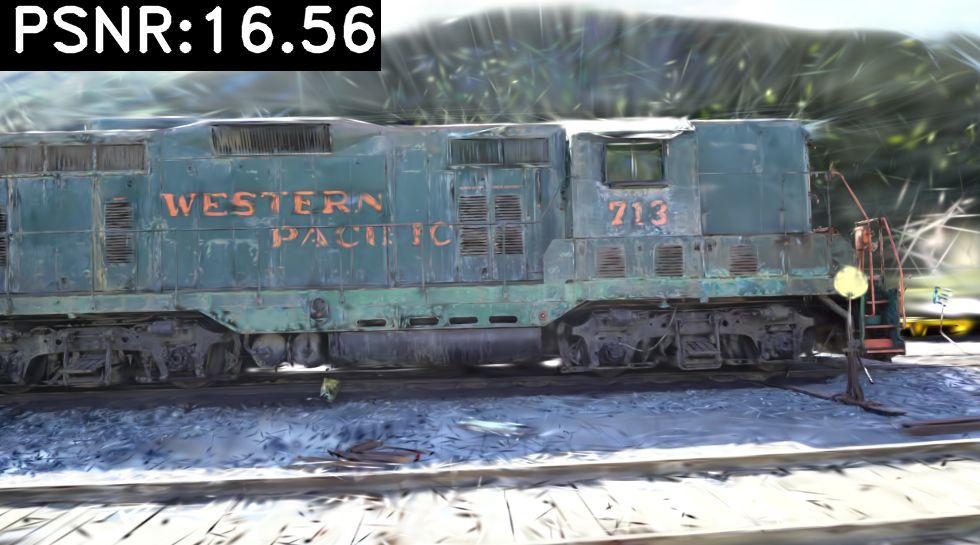}
        \caption{Base}
        \label{fig:base_approach}
    \end{subfigure}
    \hfill
    \begin{subfigure}[b]{0.49\linewidth}
        \centering
        \includegraphics[width=\linewidth]{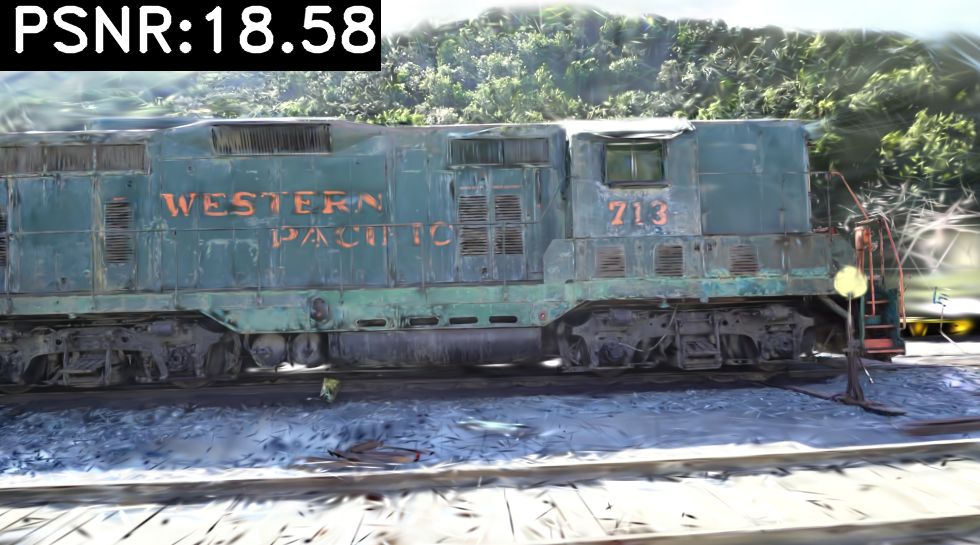}
        \caption{Base+Octree}
        \label{fig:intermediate_approach}
    \end{subfigure}
    \vspace{0.5em}
    \begin{subfigure}[b]{0.49\linewidth}
        \centering
        \includegraphics[width=\linewidth]{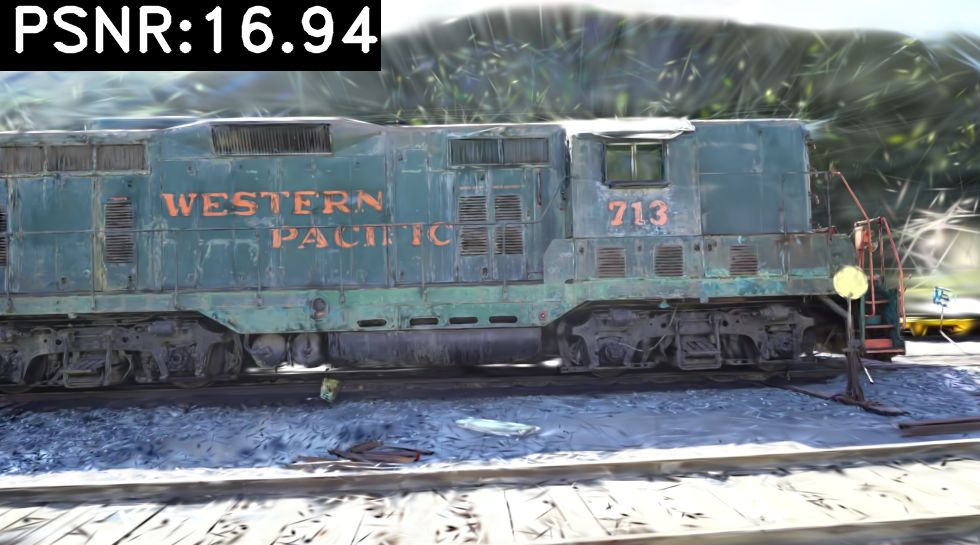}
        \caption{Base+Frustum culling}
        \label{fig:advanced_approach_c}
    \end{subfigure}
    \hfill
    \begin{subfigure}[b]{0.49\linewidth}
        \centering
        \includegraphics[width=\linewidth]{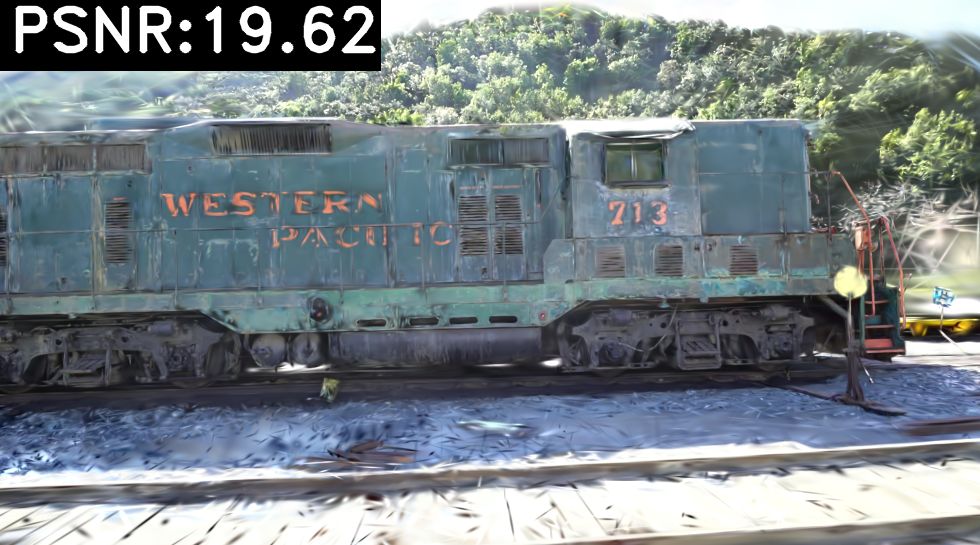}
        \caption{Base+Octree+Frustum culling}
        \label{fig:complete_approach}
    \end{subfigure}
    \caption{This figure shows the progression from our base approach to the complete approach going from top left to bottom right. Notice how the background is noticeably better when using the octree to spread out Gaussians. The impact of frustum culling becomes more noticeable when we use the octree as we enhance (b) further by forcing even more splats to lie inside the frustum.}
    \label{fig:complete_ablation_octree}
\end{figure}

The next addition is frustum culling, where we focus on loading in splats that are visible to the user. \cref{fig:advanced_approach_c} shows that the foreground becomes marginally better. As more than 10\% of all splats are visible within the current frustum, the small background splats are still not selected. This partly occurs as frustum culling occurs after the global ordering has been determined. We thus do not take into account occlusions from the current viewpoint. Using our approach with the octree results in a good representation of the scene without requiring any user-specific knowledge, while the complete approach, with frustum culling, will perform better when we have access to such information. 

\subsection{Object level}
\begin{figure}[t]
    \centering
    \begin{subfigure}{0.49\linewidth}
        \centering
        \includegraphics[width=\linewidth]{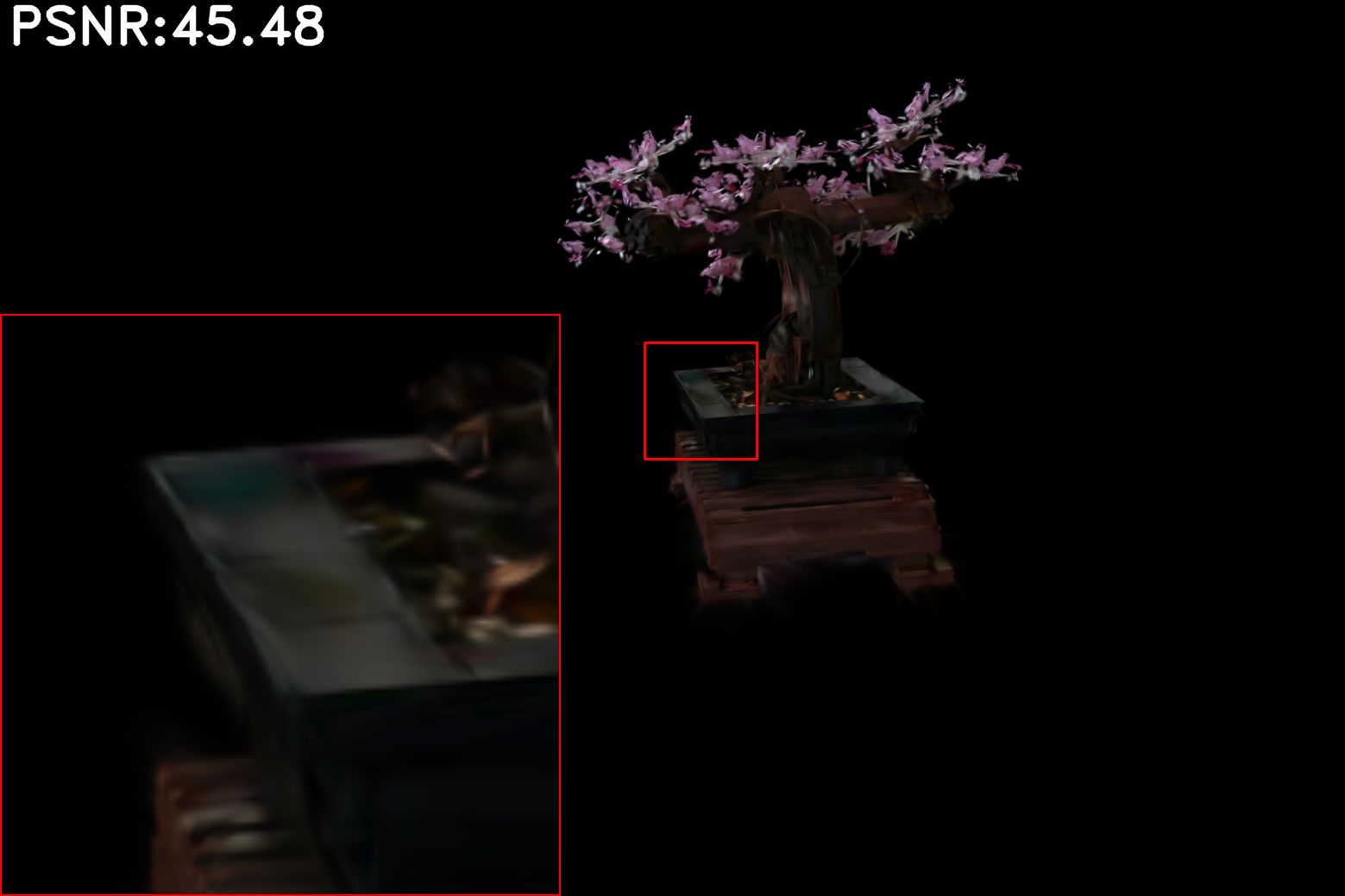}
        \caption{Ours}
        \label{fig:contr_zoomed}
    \end{subfigure}
    \hfill
    \begin{subfigure}{0.49\linewidth}
        \centering
        \includegraphics[width=\linewidth]{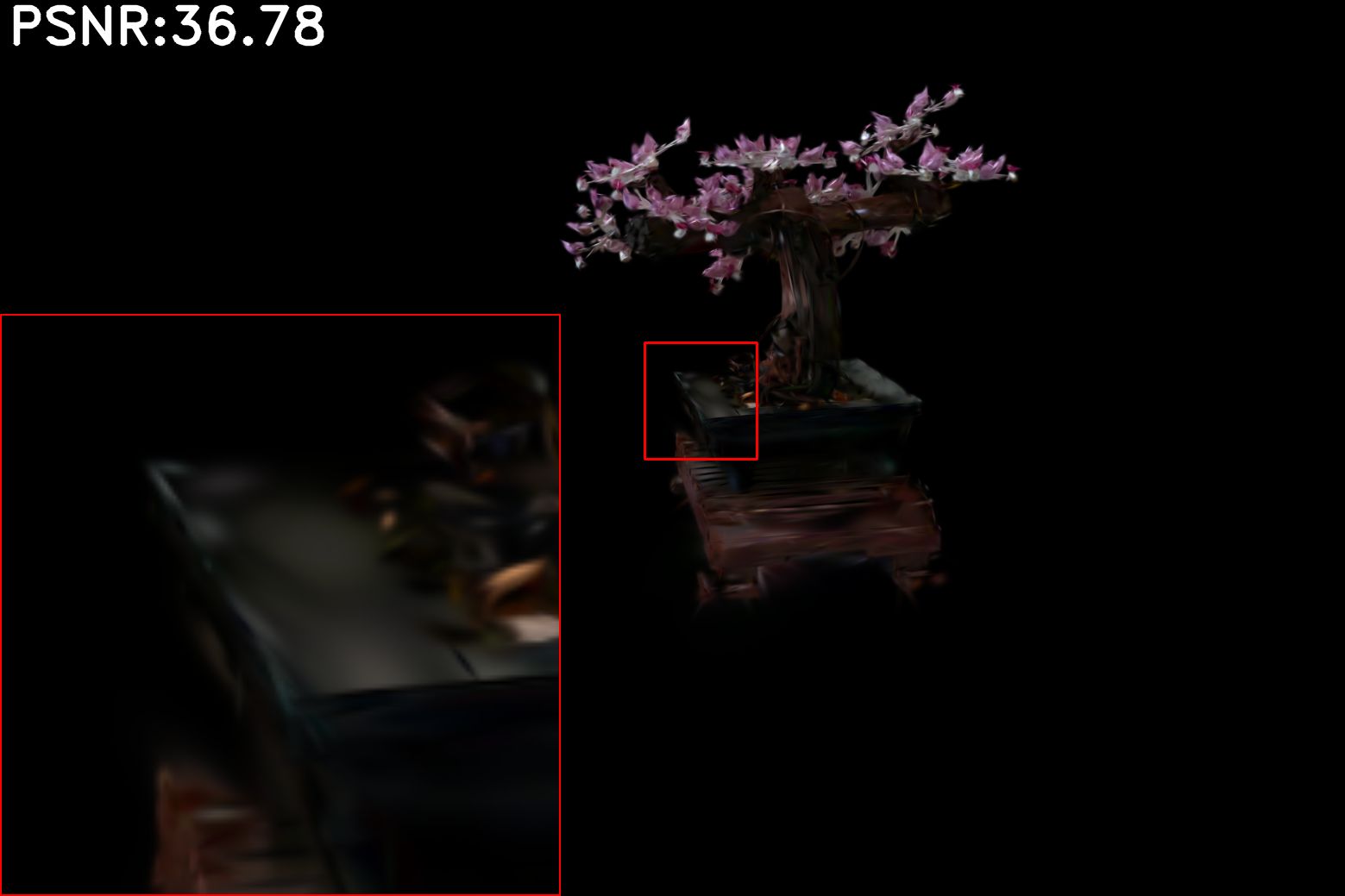}
        \caption{Antimatter}
        \label{fig:antimatter_zoomed}
    \end{subfigure}
    \caption{Comparison of our approach vs Antimatter on a segmentation of the bonsai tree from the \textit{bonsai} scene. The zoomed-in crop shows that our approach is able to load in a more detailed version, which is also reflected by the masked PSNR.}
    \label{fig:object_level}
\end{figure}

During our experiments, we manually segmented several objects and verified our approach on each object. \cref{fig:object_level} illustrates that our approach allows the progressive rendering of individual objects with each object being loaded in using solely contribution without further additions. We tested this on several objects segmented from the aforementioned datasets and consistently achieve better scores per percentage loaded splats compared to other methods. 


\subsection{Integration with compression}
\begin{figure}[h!]
    \centering
    \begin{subfigure}[b]{0.48\linewidth}
        \centering
        \includegraphics[width=\linewidth]{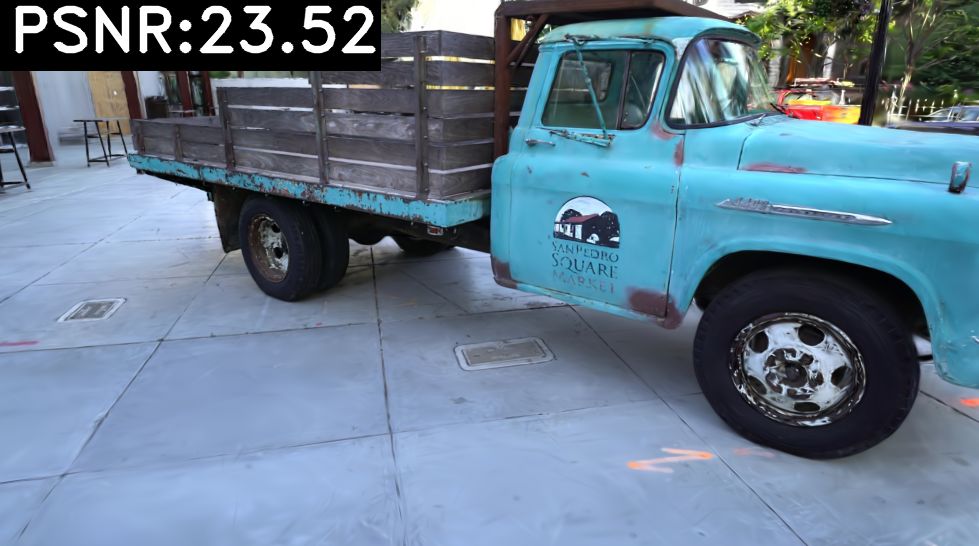}
        \caption{Ours+Compression}
        \label{fig:compr_train_contr_with_psnr}
    \end{subfigure}
    \hfill
    \begin{subfigure}[b]{0.48\linewidth}
        \centering
        \includegraphics[width=\linewidth]{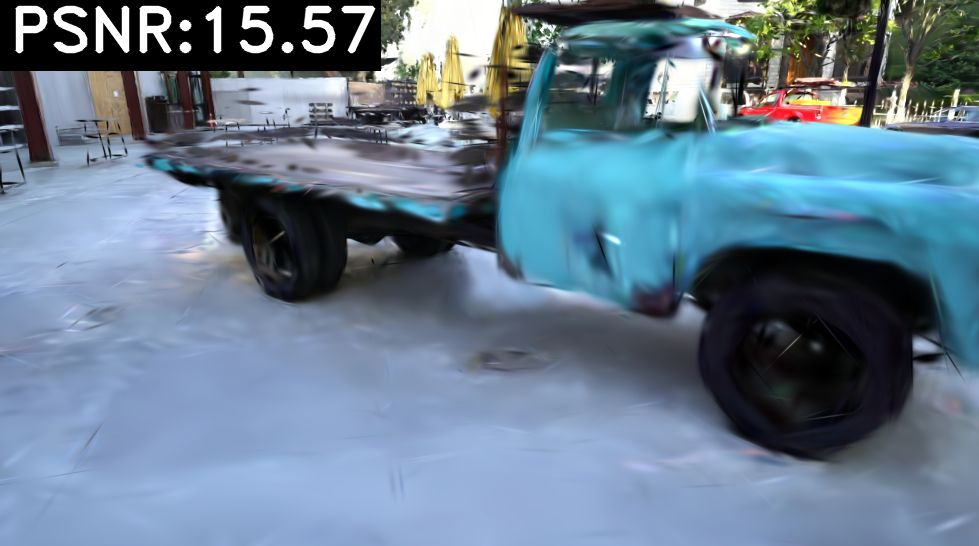}
        \caption{Antimatter+Compression}
        \label{fig:train_antimatter_with_psnr}
    \end{subfigure}
    \caption{Comparison of the compressed versions of our approach (left) and Antimatter (right), both containing only 10\% of the splats after compression.}
    \label{fig:compression_comparison}
\end{figure}

Our final contribution is the integration of our approach into an existing compression method to show the impact it can have on the quality at the time to First Paint. We adapted the technique of Niedermayr et al.~\cite{niedermayr2023compressed} to include our contribution-based ordering technique. To integrate our approach, the save-functionality has to be slightly rewritten to consider alterations from the base approach. Concretely, to perform vector quantization, the minimum and maximum values are needed for all values within said chunk. If every chunk needs to be rendered independently from the other chunks, the minimum and maximum need to be included in every chunk. Furthermore, the Gaussians are sorted in Morton order in the original approach to allow for further compression. This step happens right before saving the scene, requiring us to apply this sorting on the selected indices as well. We disabled this for now as this has no impact on the selected splats and only on the final storage requirements. \cref{fig:compression_comparison} shows that we achieve similar results comparing quality per percentage of splats compared to our base approach on the standard 3DGS implementation. In actually required memory, there is a smaller difference as the shared codebook reduces the amount of Gaussian-specific data.


\section{Conclusion and Future Work}
\label{sec:conclusion}
We proposed a novel approach that facilitates progressive rendering of 3DGS and 3DGS-inspired methods by utilizing the contribution of Gaussians to determine an order of importance among them. Our approach allows us to provide a significantly better visualization with the same amount of splats. We have also shown that our approach can work in tandem with existing compression techniques to reduce the required bandwidth further. Our approach serves as the groundwork for the progressive streaming of 3DGS, which will play a vital role in bringing 3DGS content to end-users. In the future, we can explore other perspectives to optimize the streaming order by utilizing reinforcement learning, as shown in~\cite{app132111697}. Other potential research directions include measuring user experience and application-guided progressive rendering, i.e., using object-level ordering to achieve adaptive streaming. When the domain evolves to work with larger and larger areas, progressive rendering and streaming will become more and more important.

\clearpage
\section{Acknowledgements}
This research was partly funded by the specialized FWO fellowship grant (1SHDZ24N), the European Union (HORIZON MAX-R,
Mixed Augmented and Extended Reality Media Pipeline, 101070072), the Flanders 20 Make’s XRTwin SBO project (R-12528) and the Special Research Fund (BOF) of Hasselt
University (R-14360). This work was made possible with support from MAXVR-INFRA, a scalable and flexible infrastructure that facilitates the transition to digital-physical work environments.


{\small
\bibliographystyle{ieee_fullname}
\bibliography{egbib}

\begin{thebibliography}{10}\itemsep=-1pt

\bibitem{barron2022mipnerf360}
Jonathan~T. Barron, Ben Mildenhall, Dor Verbin, Pratul~P. Srinivasan, and Peter Hedman.
\newblock Mip-nerf 360: Unbounded anti-aliased neural radiance fields.
\newblock {\em CVPR}, 2022.

\bibitem{chen24NeRFHub}
Bo Chen, Zhisheng Yan, Bo Han, and Klara Nahrstedt.
\newblock Nerfhub: A context-aware nerf serving framework for mobile immersive applications.
\newblock In {\em Proceedings of the 22nd Annual International Conference on Mobile Systems, Applications and Services}, MOBISYS '24, page 85–98, New York, NY, USA, 2024. Association for Computing Machinery.

\bibitem{chen2023neural}
Yun-Chun Chen, Vladimir~G. Kim, Noam Aigerman, and Alec Jacobson.
\newblock Neural progressive meshes, 2023.

\bibitem{cho22streamableNeRF}
Junwoo Cho, Seungtae Nam, Daniel Rho, Jong~Hwan Ko, and Eunbyung Park.
\newblock Streamable neural fields.
\newblock In {\em Computer Vision -- ECCV 2022}, pages 595--612, Cham, 2022. Springer Nature Switzerland.

\bibitem{fan2023lightgaussian}
Zhiwen Fan, Kevin Wang, Kairun Wen, Zehao Zhu, Dejia Xu, and Zhangyang Wang.
\newblock Lightgaussian: Unbounded 3d gaussian compression with 15x reduction and 200+ fps, 2023.

\bibitem{farrugia23framework}
Jean-Philippe Farrugia, Luc Billaud, and Guillaume Lavou\'{e}.
\newblock Adaptive streaming of 3d content for web-based virtual reality: an open-source prototype including several metrics and strategies.
\newblock In {\em Proceedings of ACM Multimedia Systems Conference}, MMSys '23, June 2023.

\bibitem{forgione18dash3D}
Thomas Forgione, Axel Carlier, G\'{e}raldine Morin, Wei~Tsang Ooi, Vincent Charvillat, and Praveen~Kumar Yadav.
\newblock Dash for 3d networked virtual environment.
\newblock In {\em Proceedings of the 26th ACM International Conference on Multimedia}, MM ’18, page 1910–1918, New York, NY, USA, October 2018. Association for Computing Machinery.

\bibitem{girish2024eaglesefficientaccelerated3d}
Sharath Girish, Kamal Gupta, and Abhinav Shrivastava.
\newblock Eagles: Efficient accelerated 3d gaussians with lightweight encodings, 2024.

\bibitem{DeepBlending2018}
Peter Hedman, Julien Philip, True Price, Jan-Michael Frahm, George Drettakis, and Gabriel Brostow.
\newblock Deep blending for free-viewpoint image-based rendering.
\newblock 37(6):257:1--257:15, 2018.

\bibitem{10.1145/258734.258843}
Hugues Hoppe.
\newblock View-dependent refinement of progressive meshes.
\newblock In {\em Proceedings of the 24th Annual Conference on Computer Graphics and Interactive Techniques}, SIGGRAPH '97, page 189–198, USA, 1997. ACM Press/Addison-Wesley Publishing Co.

\bibitem{10.1145/3596711.3596725}
Hugues Hoppe.
\newblock {\em Progressive Meshes}.
\newblock Association for Computing Machinery, New York, NY, USA, 1 edition, 2023.

\bibitem{hu23lfvideoGNN}
Xinjue Hu, Yuxuan Pan, Yumei Wang, Lin Zhang, and Shervin Shirmohammadi.
\newblock Multiple description coding for best-effort delivery of light field video using gnn-based compression.
\newblock {\em IEEE Transactions on Multimedia}, 25:690--705, 2023.

\bibitem{isoiec22mpegdash}
{ISO/IEC 23009-1}.
\newblock {Information technology -- Dynamic adaptive streaming over HTTP (DASH) -- Part 1: Media presentation description and segment formats}, {2022}.

\bibitem{isoiec23vpcc}
{ISO/IEC 23090-5}.
\newblock {Information technology -- Coded representation of immersive media -- Part 5: Visual volumetric video-based coding (V3C) and video-based point cloud compression (V-PCC)}, {2023}.

\bibitem{kara18evaluate_dynamic_adaptive_lf}
Peter~A. Kara, Aron Cserkaszky, Maria~G. Martini, Attila Barsi, Laszlo Bokor, and Tibor Balogh.
\newblock {Evaluation of the concept of dynamic adaptive streaming of light field video}.
\newblock {\em IEEE Transactions on Broadcasting}, 64(2):407--421, 2018.

\bibitem{kerbl3Dgaussians}
Bernhard Kerbl, Georgios Kopanas, Thomas Leimk{\"u}hler, and George Drettakis.
\newblock 3d gaussian splatting for real-time radiance field rendering.
\newblock {\em ACM Transactions on Graphics}, 42(4), July 2023.

\bibitem{Knapitsch2017}
Arno Knapitsch, Jaesik Park, Qian-Yi Zhou, and Vladlen Koltun.
\newblock Tanks and temples: Benchmarking large-scale scene reconstruction.
\newblock {\em ACM Transactions on Graphics}, 36(4), 2017.

\bibitem{lemoine23gltfStreaming}
Wouter LEMOINE and Maarten WIJNANTS.
\newblock Progressive network streaming of textured meshes in the binary gltf 2.0 format.
\newblock In {\em Proceedings of the 28th International ACM Conference on 3D Web Technology}, pages 1--11, New York, NY, USA, 2023. Association for Computing Machinery.
\newblock The 28th International Conference on 3D Web Technology.

\bibitem{lievens21RelevanceABR}
Hendrik Lievens, Maarten Wijnants, Mike Vandersanden, Peter Quax, and Wim Lamotte.
\newblock Adaptive web-based vr streaming of multi-lod 3d scenes via author-provided relevance scores.
\newblock In {\em Proceedings of the IEEE Conference on Virtual Reality and 3D User Interfaces}, VR '21, pages 488--489, 2021.

\bibitem{lievens21SLFInWeb}
Hendrik Lievens, Maarten Wijnants, Brent Zoomers, Jeroen Put, Nick Michiels, Peter Quax, and Wim Lamotte.
\newblock Adaptive streaming and rendering of static light fields in the web browser.
\newblock In {\em 2021 International Conference on 3D Immersion (IC3D)}, pages 1--8, 2021.

\bibitem{liu2021editing}
Steven Liu, Xiuming Zhang, Zhoutong Zhang, Richard Zhang, Jun-Yan Zhu, and Bryan Russell.
\newblock Editing conditional radiance fields.
\newblock In {\em Proceedings of the International Conference on Computer Vision (ICCV)}, 2021.

\bibitem{mildenhall2020nerf}
Ben Mildenhall, Pratul~P. Srinivasan, Matthew Tancik, Jonathan~T. Barron, Ravi Ramamoorthi, and Ren Ng.
\newblock Nerf: Representing scenes as neural radiance fields for view synthesis.
\newblock In {\em ECCV}, 2020.

\bibitem{mueller2022instant}
Thomas M\"uller, Alex Evans, Christoph Schied, and Alexander Keller.
\newblock Instant neural graphics primitives with a multiresolution hash encoding.
\newblock {\em ACM Trans. Graph.}, 41(4):102:1--102:15, July 2022.

\bibitem{navaneet2023compact3d}
KL Navaneet, Kossar~Pourahmadi Meibodi, Soroush~Abbasi Koohpayegani, and Hamed Pirsiavash.
\newblock Compact3d: Compressing gaussian splat radiance field models with vector quantization.
\newblock {\em arXiv preprint arXiv:2311.18159}, 2023.

\bibitem{niedermayr2023compressed}
Simon Niedermayr, Josef Stumpfegger, and Rüdiger Westermann.
\newblock Compressed 3d gaussian splatting for accelerated novel view synthesis, 2023.

\bibitem{papantonakis2024rmfsplat}
Panagiotis Papantonakis, Georgios Kopanas, Bernhard Kerbl, Alexandre Lanvin, and George Drettakis.
\newblock Reducing the memory footprint of 3d gaussian splatting.
\newblock {\em Proceedings of the ACM on Computer Graphics and Interactive Techniques}, 7(1), May 2024.

\bibitem{rakotosaona2023nerfmeshing}
Marie-Julie Rakotosaona, Fabian Manhardt, Diego~Martin Arroyo, Michael Niemeyer, Abhijit Kundu, and Federico Tombari.
\newblock Nerfmeshing: Distilling neural radiance fields into geometrically-accurate 3d meshes, 2023.

\bibitem{ravi2024sam2}
Nikhila Ravi, Valentin Gabeur, Yuan-Ting Hu, Ronghang Hu, Chaitanya Ryali, Tengyu Ma, Haitham Khedr, Roman R{\"a}dle, Chloe Rolland, Laura Gustafson, Eric Mintun, Junting Pan, Kalyan~Vasudev Alwala, Nicolas Carion, Chao-Yuan Wu, Ross Girshick, Piotr Doll{\'a}r, and Christoph Feichtenhofer.
\newblock Sam 2: Segment anything in images and videos.
\newblock {\em arXiv preprint arXiv:2408.00714}, 2024.

\bibitem{pcprog}
Markus Schütz, Gottfried Mandlburger, Johannes Otepka, and Michael Wimmer.
\newblock Progressive real‐time rendering of one billion points without hierarchical acceleration structures.
\newblock {\em Computer Graphics Forum}, 39:51--64, 07 2020.

\bibitem{slocum21via}
Carter Slocum, Jingwen Huang, and Jiasi Chen.
\newblock Via: Visibility-aware web-based virtual reality.
\newblock In {\em Proceedings of the 26th International Conference on 3D Web Technology}, Web3D '21. Association for Computing Machinery, 2021.

\bibitem{sodagar11mpegdash}
Iraj Sodagar.
\newblock The {MPEG-DASH} standard for multimedia streaming over the internet.
\newblock {\em IEEE MultiMedia}, 18(4):62--67, April 2011.

\bibitem{app132111697}
Naima Souane, Malika Bourenane, and Yassine Douga.
\newblock Deep reinforcement learning-based approach for video streaming: Dynamic adaptive video streaming over http.
\newblock {\em Applied Sciences}, 13(21), 2023.

\bibitem{vanderhooft19has_pcc}
Jeroen {Van Der Hooft}, Tim Wauters, Filip {De Turck}, Christian Timmerer, and Hermann Hellwagner.
\newblock {Towards 6DoF HTTP Adaptive Streaming Through Point Cloud Compression}.
\newblock In {\em Proceedings of the 27th ACM International Conference on Multimedia}, MM '19. Association for Computing Machinery, October 2019.

\bibitem{wijnants18HASSLF}
Maarten Wijnants, Hendrik Lievens, Nick Michiels, Jeroen Put, Peter Quax, and Wim Lamotte.
\newblock {Standards-Compliant HTTP Adaptive Streaming of Static Light Fields}.
\newblock In {\em Proceedings of the 24th ACM Symposium on Virtual Reality Software and Technology}, VRST '18, 2018.

\bibitem{xiangli2022bungeenerf}
Yuanbo Xiangli, Linning Xu, Xingang Pan, Nanxuan Zhao, Anyi Rao, Christian Theobalt, Bo Dai, and Dahua Lin.
\newblock Bungeenerf: Progressive neural radiance field for extreme multi-scale scene rendering.
\newblock In Shai Avidan, Gabriel Brostow, Moustapha Ciss{\'e}, Giovanni~Maria Farinella, and Tal Hassner, editors, {\em Computer Vision -- ECCV 2022}, pages 106--122, Cham, 2022. Springer Nature Switzerland.

\bibitem{zampoglou16adaptive_web3d}
Markos Zampoglou, Kostas Kapetanakis, Andreas Stamoulias, Athanasios~G. Malamos, and Spyros Panagiotakis.
\newblock {Adaptive Streaming of Complex Web 3D Scenes based on the MPEG-DASH Standard}.
\newblock {\em Multimedia Tools and Applications}, 77(1):125--148, Jan 2018.

\end{thebibliography}
}

\end{document}